\documentclass[eat,twocolumn]{jmlr}

\usepackage{longtable}

\usepackage{booktabs}

\usepackage{amsmath, amssymb, dsfont}

\usepackage{array,booktabs,multirow,colortbl,hhline}
\newcommand{\cell}[2]{\setlength{\tabcolsep}{0pt}\begin{tabular}{#1}#2 \end{tabular}}

\theorembodyfont{\upshape}
\theoremheaderfont{\scshape}
\theorempostheader{:}
\theoremsep{\newline}

\jmlrvolume{193}
\firstpageno{1}

\jmlryear{2022}
\jmlrworkshop{Machine Learning for Health (ML4H) 2022}

\title[Actionable Recourse via GANs for Mobile Health]{Actionable Recourse via GANs for Mobile Health}

\author{
\Name{Jennifer Chien} \Email{jennifer@benshi.ai, jjchien@eng.ucsd.edu}\\
\addr benshi.ai, UC San Diego
\AND
   \Name{Anna Guitart} \Email{guitart@benshi.ai}\\
   \Name{Ana {Fernández del Río}} \Email{ana@benshi.ai}\\
   \Name{{\'A}frica Peri\'añez} \Email{africa@benshi.ai}\\
   \addr benshi.ai
\AND
   \Name{Lauren Bellhouse} \Email{lauren@maternity.dk}\\
  \addr Maternity Foundation
 }

\newcommand\theapp{Safe Delivery App}
\newcommand\theorg{Maternity Foundation}
\newcommand\citetheapp{\citep{sda}}
\newcommand\citetheorg{\citep{mf}}
\newcommand\theotherorg{, the University of Copenhagen, and the University of Southern Denmark}

\begin{document}
\maketitle
\begin{abstract}
Mobile health apps provide a unique means of collecting data that can be used to deliver adaptive interventions\citep{marsch2021digital, buckeridge2020precision, overdijkink2018usability, dolley2018big,dowell2016four}.The predicted outcomes considerably influence the selection of such interventions. Recourse via counterfactuals provides tangible mechanisms to modify user predictions. By identifying {\it plausible} actions that increase the likelihood of a desired prediction, stakeholders are afforded agency over their predictions. Furthermore, recourse mechanisms enable counterfactual reasoning that can help provide insights into candidates for causal interventional features. We demonstrate the feasibility of  GAN-generated recourse for mobile health applications on  ensemble-survival-analysis-based prediction of medium-term engagement in the \theapp{}, a digital training tool for skilled birth attendants. 
\end{abstract}
\section{Introduction}
\label{sec:intro}
The optimization of mobile health (mHealth) application ecosystems has attracted considerable interest in recent years. Many such applications are being designed to support front-line healthcare workers through capacity building, provision of clinical reference guidelines and diagnostic/triage support, realization of patient tracking and reporting, and more. Additionally, these apps can help healthcare providers overcome challenges in limited resource settings\citep{overdijkink2018usability,hosny2019artificial,wahl2018artificial,forero2018application}.

In particular, mHealth apps allow for personalized, just-in-time interventions\citep{marsch2021digital, buckeridge2020precision, overdijkink2018usability, dolley2018big,dowell2016four}. 
Such interventions can have far-reaching downstream effects, leading to better practices, and ultimately, better health outcomes. 

This work examines the role of recourse via generative adversarial networks (GANs) \citep{Goodfellow2014} in mHealth adaptive interventions. Recourse on predicted outcomes serves two roles: First, recourse identifies the {\it plausible} ways in which users can change their predictions, which can allow stakeholders not only to better understand model predictions, but also to have agency over them. 
Second, recourse, as a feedback mechanism and learning opportunity, can provide insights into actionable intervention candidates that can modify both predictions and outcomes. 

We demonstrate this by using GANs to produce counterfactual estimates for survival-ensemble-based predictions of medium-term engagement with the \theapp{}\citetheapp{}  (hereafter, {\it the App}). The App is a digital training tool developed by \theorg{}\citetheorg\theotherorg{}, which contains evidence-based obstetric and newborn-related guidelines for supporting skilled birth attendants. 

This paper is organized as follows: the remainder of this section describes the intervention framework for mHealth and defines recourse and counterfactuals. Moreover, we introduce the App, briefly review related work, and highlight the key contributions of this research. Section \ref{sec:models} describes the models 
used to demonstrate the GAN-based recourse approach. Section \ref{sec:data-setup} defines the model specifications. The results and concluding remarks are presented in Sections \ref{sec:results} and \ref{sec:conclusions}, respectively.

\subsection{Predictions and MHealth Apps}
\label{sec:adaptive-interventions}

The bidirectional communication channel provided by data-centric mHealth apps can facilitate the collection of valuable user behavior and health outcomes information. These data can inform personalized interventions to increase the likelihood of a desired outcome\citep{Hosny2019,Wahl2018,OConnor2018,Marsch2021}. As such, personalized predictions can be used to determine optimal timing and behavioral interventions for users.

In this work, we use conditional survival forests (CSFs)\citep{csf} to predict user lifetime, i.e. how long time users will use an app before stopping altogether, i.e. churning. We use churn prediction to identify users to target for push-notification-based interventions for re-engagement. Within the App specifically, these interventions can aim to prioritize essential training for short-lifetime predicted users. \footnote{For additional information on the App, see Appendix \ref{theapp}.}

\subsection{Recourse and Counterfactuals}
\label{sec:recourse}

Recourse mechanisms for behavioral predictions allow stakeholders (i.e., decision-makers or users) to identify concrete actions to modify their prediction and provide users agency over such predictions\citep{wallin1992legal}.

In the context of mHealth apps, we define \emph{recourse} as the set of feature changes that result in the desired predicted outcome. Recourse allows the stakeholders to better understand the decision-making process and implement tangible actions to modify the predictions. Recourse mechanisms rely on contrastive/counterfactual explanations~\citep{Byrne2019, Artelt2019, Zhang2019, Klaise2021}, which inform a user \textit{why} and \textit{how} a decision was made. Moreover, recourse can be generated directly from observational data, which is particularly useful in settings in which real-world experiments are prohibitively expensive, unethical, or infeasible.\footnote{For extended discussion on recourse characteristics, see Appendix \ref{apd:recourse characteristics}.}

\subsection{Related Work}
\label{sec:related-work} 

The studies most closely related to the present research objective are \citet{Nemirovsky2020a, Nemirovsky2021ProvidingAF}, which describe recourse via GANs in the context of image generation, hiring, diabetes, and recidivism. Other studies focused on recourse generation include \citet{Wachter2017, Ustun_2019, Karimi2020}. 

In this work, recourse is studied in the context of survival-ensemble-based predictions of the time to the event of interest (here, churn). This approach to churn prediction has been described in \citet{perianez2016, bertens2017, cig2018competition, chen2019competition, Olaniyi2022}.

The App data have been previously analyzed in the contexts of content demand prediction \citep{guitart2021}, content recommendation \citep{guitart2021b}, and engagement analysis \citep{Olaniyi2022}. Moreover, the App's impact has been analyzed\citep{Lund2016, Oladeji2022}.

Furthermore, a platform to deliver adaptive interventions in mHealth solutions has been established\citep{Tang2021}. Artificial intelligence and machine learning applications in healthcare have been summarized in \citet{Davenport94}, and such applications in the global health context have been examined in \citet{Hosny2019, Wahl2018}.

\subsection{Our Contribution}
\label{sec:our-contribution}

To the best of our knowledge, this work represents the first attempt at applying a GAN-based recourse approach to survival analysis and predictions in the context of mHealth interventions. The results demonstrate the two primary impacts of providing recourse: identification of candidate interventions and resulting prediction modification within an mHealth intervention framework.


\section{Methods and Models}
\label{sec:models}
We describe the methods and models used, specifically, CSFs and their use in churn prediction (Appendix \ref{sec:csf}) and recourse via GANs and the counterfactual estimation it enables (Section \ref{sec:gans}).

\subsection{GANs and Counterfactual Reasoning}
\label{sec:gans}

GANs \citep{Goodfellow2014} are composed of two artificial neural networks: a generator that produces realistic synthetic data, and a discriminator that differentiates between the generator output and real data. A GAN is trained through an adversarial min-max game in which the generator and discriminator are alternately trained. GANs can provide actionable, realistic, fast actions by a single feed-forward pass through a neural network.

GANs directly learn the data manifold from the training data, which is advantageous for recourse: First, the GAN output is realistic because the pre- and post-recourse action data are similar to the data manifold. Second, model owners can use GANs as a model of the data manifold to simulate interventional distributions to narrow the solution space for interventions and identify important actions.

CounteRGANs \citep{Nemirovsky2020a} generate counterfactual reasoning via feasible changes that result in a specified classification. A CounteRGAN is a specialized GAN composed of a generator $G$, discriminator $D$, and fixed classifier $C$. We define the features of user $i$ as $x_i$, with the corresponding predicted and true labels being $C(x_i)$ and $y(x_i) \in \{0, 1\}$, respectively, denoting $y=1$ as the desired outcome. $D$ is trained on data from the underlying true data distribution $p_{data}$. In contrast, $G$ is trained only on data $x_i \; \forall i \textnormal{ s.t. } y(x_i) = 0$, i.e., for all users that churn. The output of $G(x_i) = a_i$ is the delta change that when added to the user's original feature results in the desired predicted class, $C(x_i + a_i) = 1$.

\section{Model Specifications and Data}
\label{sec:data-setup}

We use data sampled from the App users from January 1, 2018 and October 1, 2021 from countries with the maximum App use (India, Myanmar, Ethiopia...). Further information on feature construction is in Appendix \ref{apd:app data}.

\begin{figure*}[ht]
    \centering
    \resizebox{0.8\linewidth}{!}{
    \includegraphics{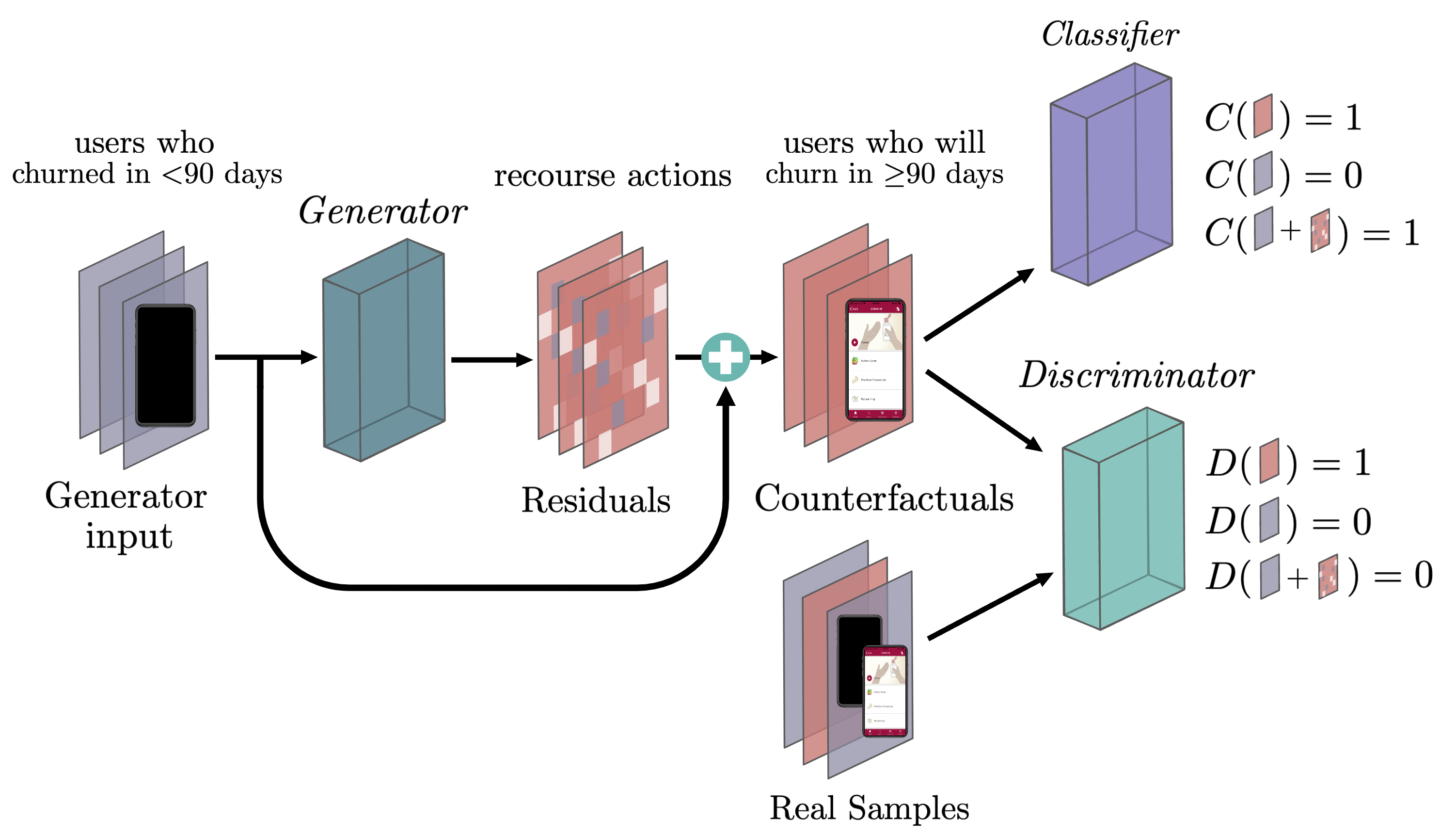}
    }
    \caption{CounteRGAN Architecture for the App Data. Two neural networks are used: generator $G$ trained to output residuals, and discriminator $D$ trained to distinguish real and augmented data. Gray tiles labeled ``generator input'' represent data for users that churned in $<$90 days ($y(x_i)=0$). The red tiles labeled ``counterfactuals'' represent post-recourse users that are predicted to use the App for $\geq$90 days ($C(x_i+a_i)=1$). The patchwork tiles represent residuals $a_i$. The gray and red stacked tiles labeled ``real samples'' represent the training data $y(x_i)=1$.
    }
    \label{Fig::architecture}
\end{figure*}

The model setup is shown in Figure \ref{Fig::architecture}. We examine GAN-generated recourse for users predicted to engage with the App for less than 90 days (period considered a marker of medium-term engagement). Further information on CSF and CounteRGAN training is in Appendix \ref{model training}.

We train CounteRGANs with survival forests involving 1, 5, and 20 trees, henceforth referred to as \textit{model}\_\{\textit{number of trees}\}. We train CounteRGAN on various sizes of CSFs to determine the influence of the model performance on recourse effectiveness. In addition, we generate counterfactuals with regularized gradient descent (RGD) implemented in Alibi, as described in \citep{Klaise2021} with the 20-tree survival forest as a baseline.

\section{Results}
\label{sec:results}

We describe the observed model performance and time requirements in Section \ref{Sec::results::modelperformance}, and Section \ref{Sec::results::efficacyofrecourse} discusses the efficacy and cost of recourse. For concreteness, we provide an example of required feature changes to achieve successful recourse to a single user in Appendix \ref{recourse action set example}.

\subsection{Model Performance and Time Requirements}
\label{Sec::results::modelperformance}
As expected, model accuracy increases with the forest size for both the initial and post-recourse classifier. 
In addition, we measure mean clock time to compute the recourse(in seconds), finding that recourse generation via GANs is faster than that based on RGD, as it only requires a single feed-forward pass through the generator network to provide personalized recourse (instead of solving an optimization problem per user). Additional details are presented in the appendix \ref{app}.

\begin{figure*}[htb]
    \centering
    \resizebox{0.60\linewidth}{!}{
    \includegraphics{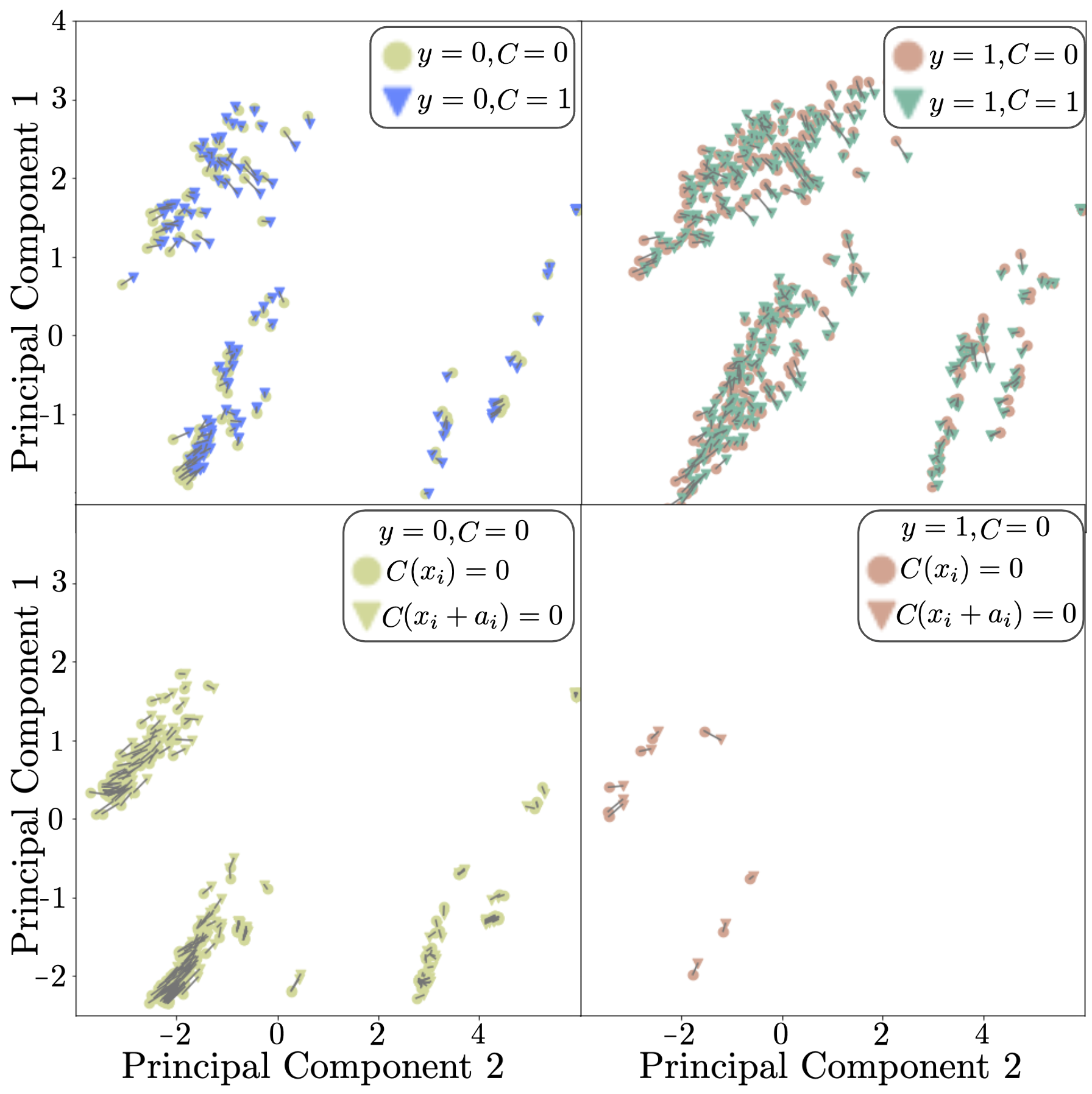}
    }
    \caption{Scatterplot of Pre-Recourse (circles) and Post-Recourse (triangles) User Features. The original training data are transformed via principal component analysis for ease of visualization. Users are separated by predicted and true outcomes: $y=0$ and $y=1$ on the left and right columns, respectively; effective recourse ($C=1$) and ineffective recourse ($C=0$) are on the top and bottom halves, respectively. }
    \label{Fig::sidebyside_scatter}
\end{figure*}

\subsection{Efficacy and Cost of Recourse}
\label{Sec::results::efficacyofrecourse}

In this section, we examine the efficacy and cost of recourse as an informational tool to the range of feasible and infeasible actions. In addition, we demonstrate how recourse can be used as an auditing mechanism for identifying subgroups for performance disparities, as recourse is dependent on model predictions.

To quantify the efficacy of recourse via CounteRGANs, we measure the percentage of users provided effective recourse ${C(x_i+a_i)=1}$ and denied recourse ${C(x_i+a_i)=0}$. We find that the model\_20 CounteRGAN provides the highest percentage of effective recourse with 29.3\% compared to 15.6\% with the model\_20 RGD. See additional details in Appendix \ref{apd:recourse efficacy}.

Differences in the efficacy of recourse can serve as a dataset- and model-specific auditing mechanism. For instance, Figure \ref{Fig::sidebyside_scatter} shows the pre- and post-recourse user features, separated by the true outcome ($y=0$ and $y=1$ for the left and right columns, respectively) and recourse efficacy (effective and ineffective recourse in the top and bottom rows, respectively). For ease of visualization, we show the features reconstructed under the top two components using principal component analysis \citep{Pearson1901}. 

Two observations can be made: First, across users with ineffective recourse (bottom row of Figure \ref{Fig::sidebyside_scatter}), the $y=1$ individuals are confined to a narrower range of principal component 2 than those with $y=0$.
This finding can inform the range of infeasible recourse actions.
Similarly, a comparison of the recourse efficacy (top and bottom rows in Figure \ref{Fig::sidebyside_scatter}) shows that the recourse efficacy may be higher with larger values of principal component 1.
Although these results are dataset specific, they can highlight how recourse computation and analysis can clarify the domain of feasible actions.

Second, we observe overlap in the regions occupied by both effective and ineffective recourse. Although this result indicates that the recourse actions being provided are realistic and feasible, it may also indicate that a one-size-fits-all approach to providing recourse is likely insufficient. There may be user subgroups to which we cannot provide effective recourse via GANs. These subgroups may require additional incentives or indicate model performance disparities. 

Thus, even in the case of ineffective recourse, stakeholders can still learn from group differences. Recourse can be used as an auditing mechanism aimed at answering the following questions:
Can we provide users with recourse? Which groups of users cannot be provided with effective recourse? Is the model performance inferior for this subgroup? For this subgroup, does the GAN require better data or do we need to train a different model for recourse estimation?
%
%


We provide further analysis and discussion of the cost of recourse in Appendix \ref{apd:cost of recourse}.



\section{Summary and Conclusions}
\label{sec:conclusions}

We develop a GAN-based approach to provide recourse on survival-forest-based predictions and demonstrate its application to obtain medium-term engagement predictions for users of the \theapp{}. The findings highlight the potential of obtaining fast, effective recourse via GANs in mHealth applications, its use as an auditing mechanism to locate disparities in recourse feasibility across model performance and user groups, and its role as an informational agent to audit misalignment between predicted and real-world feature importance. 

The proposed approach provides a pipeline for identifying intervention-worthy features and causal recourse action candidates, drawing inferences directly and solely out of observations. This approach can provide all stakeholders with greater understanding and agency over model predictions (and hence, prediction mediated interventions) and mechanisms affecting the actual outcomes that are being predicted, enabling the definition of additional interventions. 

In this context, recourse via GANs is likely to emerge as a key component of trustworthy adaptive interventions for mobile health toolkits. 

\vspace{-0.2in}

\acks{The authors thank Archana Choudhary for reviewing the manuscript. This work was supported, in whole or in part, by the Bill \& Melinda Gates Foundation INV-022480. Under the grant conditions of the Foundation, a Creative Commons Attribution 4.0 Generic License has already been assigned to the Author Accepted Manuscript version that might arise from this submission.}

\bibliography{chien22}


\appendix

\section{The \theapp{}}
\label{theapp}

Maternal and neonatal mortality remains a pressing problem in global health. Each year, nearly 300,000 women and 5 million newborns die of causes directly related to pregnancy and childbirth \citep{united2020levels,unfpa2020}. In addition, nearly all newborn deaths occur in low- and middle-income countries, 80\% of which are preventable and treatable by cost-efficient interventions \citep{wh02104, Lawn2016}. 

Digital tools such as the App can help build the capacities of skilled birth attendants\citep{Lund2016,Oladeji2022}. The App is divided into clinical modules covering key topics related to maternal and newborn health, including both normal labour and birth as well as guiding healthcare workers through common obstetric and newborn complications. Each module consists of educational videos, easily referenceable action cards, drug lists, practical procedures, and a series of tests that increase in difficulty to assess the depth of knowledge and skills acquired by users.

\section{Recourse Characteristics}
\label{apd:recourse characteristics}
As defined, recourse must have the following four characteristics: First, recourse actions must be \textit{actionable}, i.e., they must use features that users can change \citep{Joshi2019, Karimi2020, Ustun_2019}. For example, recourse may be provided to a user for increasing their lifetime in the app, but not by changing the date of first use. Second, recourse actions must be \textit{realistic}~\citep{Wachter2017}, following real-world directional restrictions of features (e.g., preservation of the monotonically increasing nature of accumulated metrics). Third, recourse actions must be \textit{feasible}~\citep{Ustun_2019} and not prohibitively expensive to execute (e.g., increasing the number of modules completed weekly by a \textit{reasonable} amount). Finally, recourse actions must \textit{obtain the desired predicted outcome}, meaning that the implementation of such actions must result in the desired predicted outcome. 

\section{CSFs and Churn Prediction}
\label{sec:csf}

Survival analysis models have historically been used in medical and biological research for estimating life expectancy \citep{clark, Fleming2000, Quinlan1986}. However, the survival analysis toolbox provides methods generalizable to modeling and predicting the time to an event of interest in the presence of right censoring. CSFs \citep{csf} are a variant of the random survival forests (RSFs) \citep{Ishwaran2008} that recursively partition the feature space pertaining to the splitting criteria through linear rank statistics. This method minimizes the bias typically encountered in RSF predictions.

In this context, CSFs can be used to predict how long users will remain active before they churn (i.e., the number of days between the first and last login) \citep{perianez2016, bertens2017, cig2018competition, chen2019competition, Olaniyi2022}. We define churn as a number of consecutive days without in-App activity, that is determined following \citet{Olaniyi2022}.

\section{App Data and Feature Construction}
\label{apd:app data}
We use the data pertaining to a sample of users of the App between January 1, 2018 and October 1, 2021, from countries with the maximum App use (India, Myanmar, Ethiopia...). We use a train/test split of 50/50, resulting in datasets of 730 and 729 users, respectively. The in-App logs are processed into daily user metrics, such as the number of daily sessions, time spent using the App, days since last login, and similar values for specific e-learning contents to characterize the user activity and behavior. The metrics are used to construct features by applying statistical operations (mean, max, normalization, etc.) over different periods, resulting in 181 features. 

\section{Model Traning}
\label{model training}
First, the CSF models are trained and used to predict days to churn for the users. Next, the predicted days to churn are converted to a binary label predicting whether the user will have a lifetime of at least 90 days.

Next, we input feature data from users that churned in less than 90 days to $G$ and train it iteratively via gradient descent. 
We add the generator output and original input to construct the counterfactuals of users that achieve a predicted lifetime of at least 90 days and appear realistic to the discriminator. The discriminator is trained to distinguish users that will churn in 90 days or more and those that have implemented recourse actions to extend their predicted lifetime to at least 90 days.

Each CounteRGAN is trained for up to 600 iterations but saved at checkpoints at which sufficient discriminator performance is observed. These checkpoints correspond to an accuracy less than or equal to 0.55 for the pre- and post-recourse users. Low accuracy on pre- and post-recourse data indicate that the post-recourse data are difficult to distinguish and thus more realistic.

\section{Metric Definitions}
\label{apd: metric definitions}
The CSF and GAN performances are evaluated using the following metrics:
\begin{itemize}
    \item Accuracy: $\frac{1}{n} \sum_{i}^{n} \mathds{1}(y_i, \hat{y_i})$. 
    \begin{itemize} 
        \item Model Accuracy: $n$ is the total number of users in a dataset (training or testing, $y \in \{0, 1\}$ or $y \in \{0\}$, respectively). The accuracy indicates the model performance in terms of the number of predictions that match the true observed label.
        \item Discriminator Accuracy: This metric indicates the discriminator performance in terms of the number of predictions that match the true \emph{original data} or \emph{data augmented by the generator}.
        \item Classifier Accuracy: $n$ is the total number of $y=0$ users that are augmented sufficiently by the generator to obtain a $\hat{y} = 1$ prediction.
    \end{itemize}
    \item Percent Denied: $\frac{1}{n}\sum_{i} \mathds{1}(\hat{y}(x_i), 0)$, where $n$ is the number of applicants in the dataset (training/testing). The metric indicates the proportion of applicants predicted to churn in less than 90 days, thereby qualifying for recourse.
    \item Percent Successful Recourse: \\
    $\frac{1}{||\hat{y}(x_i) = 0||} \sum_{i \in \hat{y}(x_i) = 0} \mathds{1}(\hat{y}(x_i+a_i), 1)$. This metric indicates the proportion of applicants predicted to churn more than 90 days after adopting the recourse actions recommended by the generator.
    \item Mean Cost of Successful Recourse Actions: 
    $\frac{1}{||\hat{y}(x_i+a_i) = 1||} \sum_{i \in (\hat{y}(x_i+a_i) = 1 )} ||a_i||^2$. This metric indicates the average $L_2$ distance for the feature change for applicants who successfully achieve the desired predicted outcome.
    \item Cumulative Cost of Denied Recourse: 
    $\sum_{i \in (\hat{y}(x_i+a_i) = 0 )} ||a_i||^2$. This metric indicates the total $L_2$ distance for the feature change for applicants who do not achieve their predicted desired outcome.
    \item Average Clock Time: $\frac{1}{n}\sum_{i}T_{end}(x_i) - T_{start}(x_i)$, where $n$ is the number of users in the dataset (training/testing), and $T$ is the clock time at which the recourse calculation starts and ends. This metric indicates the average time (in seconds) required to calculate a user's recourse actions to estimate the corresponding counterfactual.
\end{itemize}

\section{Recourse Action Set Example}
\label{recourse action set example}

\begin{table*}[htbp]
    \centering
    \resizebox{\linewidth}{!}{
    \begin{tabular}{lcc}
        \toprule
        \cell{l}{Features to change} &
        \cell{l}{Original} &
        \cell{l}{Required} \\
        
        \toprule
        \textit{Number of days between engaged actions for the first 15 active days} &
        1.0 &
        1.3041 \\
        
        \textit{Action count for the last 15 active days} &
        1.0 &
        1.2347 \\
        
        \textit{Connection time for the last 60 active days} &
        1.0 &
        1.1755 \\
        
        \textit{E-learning action count for the first 30 active days} &
        1.0 &
        1.1716 \\
        
        \textit{Number of connected days for the first 15 active days} &
        1.0 &
        1.1660 \\
        \bottomrule
    \end{tabular}
    }
    \caption{Example of a recourse action. We provide the top five largest changes constituting a successful recourse action for a single user under model\_20. ``Original'' indicates the user's original feature values, and ``Required'' lists the feature values required for the users to be predicted to churn in $\geq$ 90 days. All features here are maximum normalized. In this case, the `\_norm\_\_max' extension on a feature indicates the maximum normalized}
    \label{recourse_example_table}
\end{table*}

\section{Model Performance and Time Requirements}\label{app}
The initial classifier performance over all users and those who churned in less than 90 days is presented in Table \ref{metrics_table_accuracy}. As expected, the model performance is enhanced with the forest size, from 0.7177 for all users for model\_1 to 0.8709 for model\_20. The percent of denied users is consistent across models (Table \ref{metrics_table}), increasing slightly from 44.96\% for model\_1 to 52.34\% for model\_20. The lack of significant variation is attributable to the binarization of the survival prediction. 

Low discriminator performance on both real and fake data signifies realistic ``generated'’ (transformed) data (Table \ref{metrics_table_accuracy}). In addition, for post-recourse applicants, the model performance is enhanced with the forest size. This phenomenon likely occurs because higher accuracy CSFs can provide more precise feedback to the GAN, allowing it to better learn the data manifold and differentiate classes. Therefore, model\_20 GANs represent the optimal platform.

We analyze the time required by different models to compute the recourse by comparing the mean clock time (in seconds), as indicated in Table \ref{metrics_table}. On the whole, recourse generation via GANs is faster than that based on RGD. This faster runtime is particularly salient in low- and middle-income countries, in which the internet connectivity may often be unreliable or user computational resources may be severely limited.

\begin{table*}[htb!]
    \centering
    \resizebox{\textwidth}{!}{
    \begin{tabular}{lcccc}
        \toprule
        \cell{l}{\\Counterfactual\\Model} &
        \cell{c}{Conditional\\Survival Forest\\ Num Trees} &
        \cell{c}{Initial Model\\Accuracy\\($y\in\{0, 1\}/\{0\}$)} &
        \cell{c}{Discriminator\\Accuracy\\($D(x_i)$/$D(x_i+a_i)$)} &
        \cell{c}{Post-Recourse\\Classifier\\Accuracy} \\
        \toprule
        \cell{l}{via GANs} & 
        1 & 
        \cell{l}{0.718 / 0.656} & 
        \cell{l}{0.194 / 0.806} &
        \cell{l}{0.372} \\
        
        &
        5 & 
        \cell{l}{0.813 / 0.810} & 
        \cell{l}{0.613 / 0.278} &
        \cell{l}{0.497} \\
        
        &
        20 & 
        \cell{l}{0.871 / 0.868} &
        \cell{l}{0.485 / 0.512} &
        \cell{l}{0.627} \\
        
        \midrule
        
        \cell{l}{via RGD} & 
        \cell{l}{20} & 
        \cell{l}{0.871 / 0.868} & 
        \cell{l}{-} &
        \cell{l}{0.627} \\
        
        \bottomrule
    \end{tabular}
    }
    \caption{Validation Metrics. Accuracy metrics for the proposed counterfactual models based on GANs with CSFs with varying number of trees {1, 5, 20} converted to binary classifiers. Results of a baseline model that produces counterfactuals via regularized gradient descent with a 20-tree CSF converted to a binary classifier are also presented for comparison.
    \label{metrics_table_accuracy}}
\end{table*}

\section{Efficacy of Recourse}
\label{apd:recourse efficacy}
Table \ref{metrics_table} summarizes the values of the percentage of users provided $C(x_i+a_i)=1$ and denied recourse $C(x_i+a_i)=0$ after implementing a CounteRGAN-generated action. Although the percentage of recourse-denied users is consistent across models, the percentage afforded recourse increases for larger forests. This result supports that CounteRGANs trained on more accurate models may be able to provide more effective recourse. In addition, both models\_5 and \_20 provide more effective recourse than that of counterfactuals obtained via RGD. 

\begin{table*}[htb!]
    \centering
    \resizebox{\textwidth}{!}{
    \begin{tabular}{lcccccc}
        \toprule
        \cell{l}{\\Counterfactual\\Model} &
        \cell{c}{Conditional\\Survival Forest\\ Num Trees} &
        \cell{c}{\% Denied \\ ($C=0$)} &
        \cell{c}{\% Successful \\ Recourse \\ ($C=1$)} &
        \cell{c}{Mean Cost\\of Successful\\Recourse\\Actions} & 
        \cell{c}{Cumulative\\Cost of\\Denied\\Recourse}&
        \cell{c}{Mean\\Compute\\Time\\(seconds)}\\
        \toprule
        \cell{l}{via GANs} & 
        1 & 
        \cell{l}{45.0\%} & 
        \cell{l}{4.7\%} & 
        \cell{l}{1.5496} &
        \cell{l}{319.22} &
        \cell{l}{0.0006} \\
        &
        5 & 
        \cell{l}{51.9\%} & 
        \cell{l}{16.1\%} & 
        \cell{l}{1.7873}  &
        \cell{l}{441.45} &
        \cell{l}{0.0003} \\
        
        &
        20 & 
        \cell{l}{52.3\%} & 
        \cell{l}{29.3\%} & 
        \cell{l}{1.6833}  &
        \cell{l}{535.30} &
        \cell{l}{0.0005} \\
        
        \midrule
        \cell{l}{via RGD} & 
        \cell{l}{20} & 
        \cell{l}{52.3\%} & 
        \cell{l}{15.6\%} & 
        \cell{l}{0.0702}  &
        \cell{l}{3.45} &
        \cell{l}{132.3800} \\
        
        \bottomrule
    \end{tabular}
    }
    \caption{Recourse Generation Metrics. Metrics to evaluate the recourse performance, cost, efficiency and runtime for the proposed counterfactual models and baseline counterfactual model. Results of a baseline model that produces counterfactuals via regularized gradient descent with a 20-tree CSF converted to a binary classifier are also presented for comparison.}
    \label{metrics_table}
\end{table*}

\section{Cost of Recourse}
\label{apd:cost of recourse}

\begin{figure*}[htbp]
    \centering
    \resizebox{\linewidth}{!}{
    \includegraphics{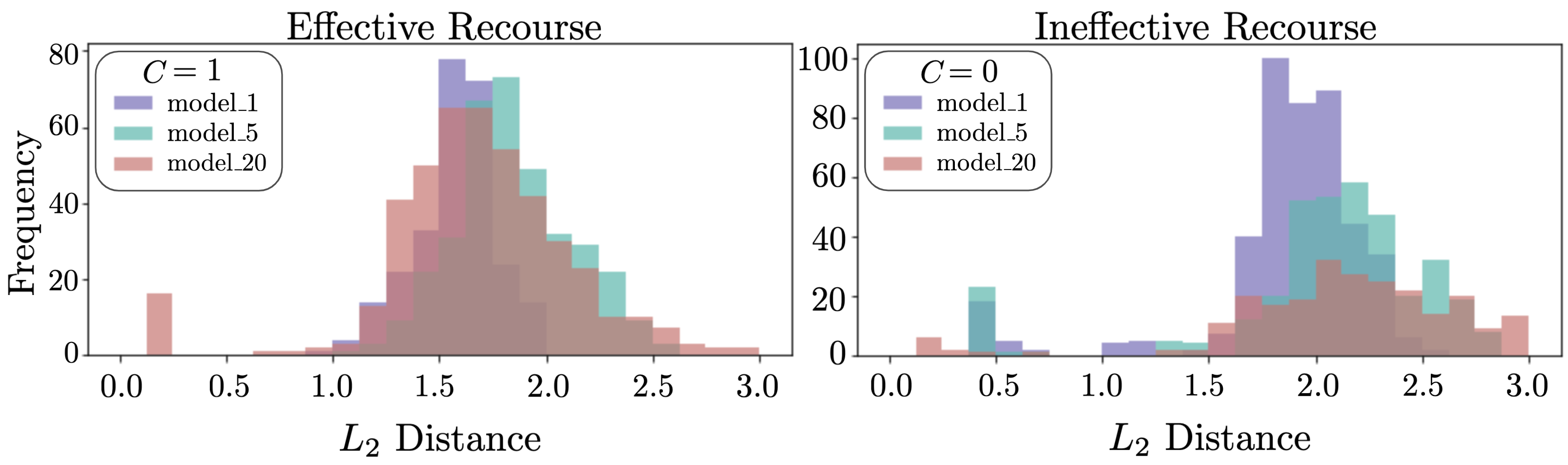}}
    \caption{Histogram of the Recourse Cost by Efficacy. Cost distribution for the different models for effective (left) and ineffective (right) recourse. For the considered dataset, model\_20 is associated with a wider range of recourse cost than model\_5 and model\_1.}
    \label{Fig::recourse_histogram}
\end{figure*}

We examine the cost of recourse as another auditing mechanism. We estimate the cost of recourse with $L_2$ distance across effective and ineffective recourse. Table \ref{metrics_table} presents the mean cost of an effective recourse action and cumulative denied recourse cost across models. Model\_1 has the lowest mean cost of successful recourse actions, followed by models\_20 and \_5. However, the cumulative denied recourse cost does not exhibit the same trend: it increases from model\_1 to models\_5 and \_20, thereby highlighting the differences in the cost of recourse across user groups. Figure \ref{Fig::recourse_histogram} shows the differences in the cost distributions of effective (left) and ineffective (right) recourse. The cost of effective recourse is similar across models. However, the cost of ineffective recourse differs across models, with model\_20 associated with a wider range but lower frequency.

This finding demonstrates that recourse may be providing disproportionate gains and losses across those who received effective and ineffective recourse. In addition, the comparison of recourse via CounteRGANs and RGDs shows that RGDs have a lower cost, in terms of both the mean cost of effective recourse action and cumulative denied cost. The effects of lower efficacy recourse must be considered as it may have negative downstream effects such as lack of accountability, lack of trust, and potential disengagement with the App.

\begin{figure*}[htb]
    \centering
    \resizebox{\linewidth}{!}{
    \includegraphics{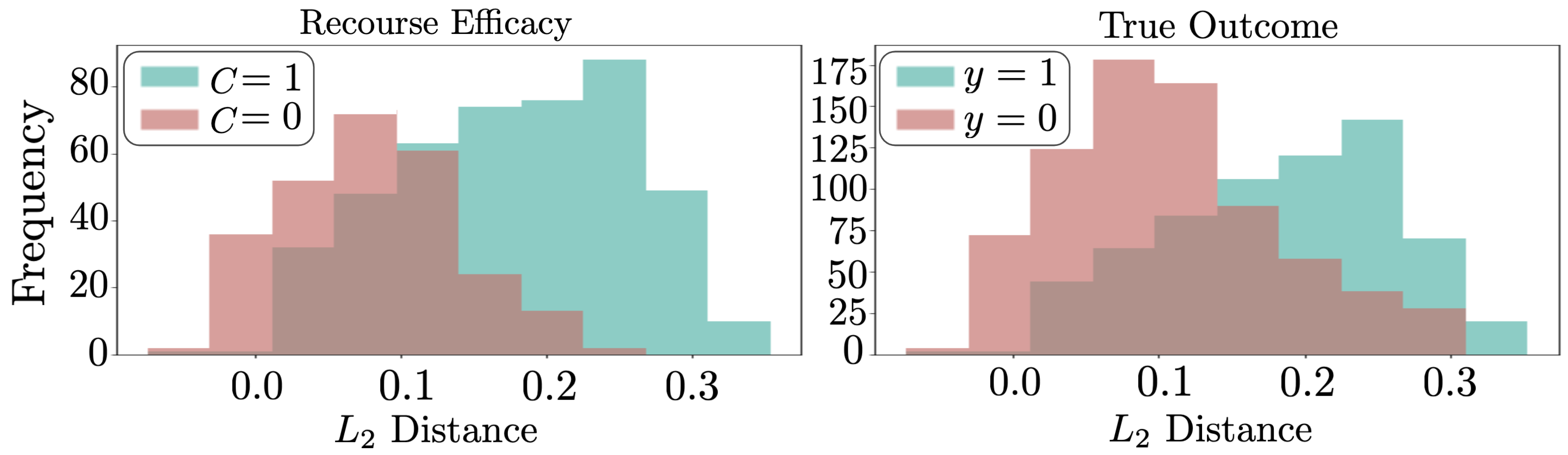}}
    \caption{Histograms of the Recourse Cost Associated With a Specific Feature Separated by Recourse Efficacy (left) and True Outcome (right). The feature is the maximum normalized number of actions taken by a user in a day over the last 15 days that the user was active. The costs correspond to model\_20. The distributions across the predicted and true outcome differ in terms of the range, i.e., $y=0$ has a wider range than $\hat{y} = 0$, indicating disparities in the estimated and true causal impacts of a feature.}
    \label{Fig::costofrecourselast15}
\end{figure*}

To demonstrate the utility of recourse as an informational tool, stakeholders can examine the distributions of the cost of recourse across true and predicted outcomes. Figure \ref{Fig::costofrecourselast15} shows two histograms of the cost of recourse associated with a particular feature: the maximum number of actions taken in the last 15 days. The plot to the left shows the distribution across effective and ineffective recourse actions. The plot to the right shows the corresponding distribution across true outcomes. 

The cost distributions differ when separated by the predicted outcome and true label. This finding indicates that this feature may not be causally related to the true outcome and may thus have an inflated estimation effect on the true outcome. Examining these distributional differences across important features can help audit the model trust, especially as it is used to provide recourse suggestions to users.

\section{Data and Code Availability}
All data used in this analysis are derived from the \theapp{} logs and belong to the Maternity Foundation. For inquiries regarding the use of these data, please contact the Maternity Foundation at mail@maternity.dk. The code used is available at https://github.com/benshi-ai/paper\_2022\_ml4h\_recourse\_gans.

All the analysis were performed with Python version 3.7.13 on Google Colab, using the following packages from PyPI: pysurvival 0.1.2 \citep{pysurvival_citep}, alibi 0.7.0 \citep{klaise2021alibi}, and tensorflow 2.8.2 \citep{tensorflow2015-whitepaper}.

\end{document}